\documentclass[11pt]{article}

\usepackage[utf8]{inputenc}
\usepackage[T1]{fontenc}
\usepackage{lmodern}
\usepackage[margin=1in]{geometry}
\usepackage{amsmath,amssymb,amsthm}
\usepackage{graphicx}
\usepackage{booktabs}
\usepackage{multirow}
\usepackage{array}
\usepackage{xcolor}
\usepackage{microtype}
\usepackage[numbers,sort&compress]{natbib}
\usepackage{pgfplots}
\pgfplotsset{compat=1.17}
\usetikzlibrary{arrows.meta,positioning,fit,backgrounds,calc}
\usepackage{algorithm}
\usepackage{algpseudocode}
\usepackage[colorlinks=true,linkcolor=blue,citecolor=blue,urlcolor=blue]{hyperref}
\usepackage{cleveref}

\newtheorem{definition}{Definition}

\newcommand{\ztil}{\tilde{z}}
\newcommand{\zadv}{\tilde{z}_{\mathrm{adv}}}
\newcommand{\zcln}{\tilde{z}_{\mathrm{clean}}}
\newcommand{\ztgt}{\tilde{z}_{\mathrm{tgt}}}
\newcommand{\Linf}{\ell_\infty}

\DeclareMathOperator*{\argmax}{arg\,max}

\title{\textbf{Attacking the Trusted Imagination:\\
Oracle-Level Integrity Attacks on Imagine-then-Act World Models}}

\author{
  Linghan Chen \\
  Adelaide University \\
  chenlinghan2004@163.com
  \and
  Kaiyan Ji \\
  Adelaide University \\
  kaiyanji56789@gmail.com
  \and
  Minyu Guo\textsuperscript{*} \\
  Adelaide University \\
  minyu.guo@adelaide.edu.au
}
\date{\today}

\begin{document}
\maketitle

\begin{abstract}
Many recent vision-language-action (VLA) policies adopt an \emph{imagine-then-act} design. A world-action model (WAM) first imagines a short future as a latent trajectory $\ztil$, on which the action is then conditioned. We identify this trusted imagination, rather than the reactive policy, as the exposed attack surface. A downstream oracle, such as a safety gate, a visual model-predictive-control (MPC) planner, or an imagine-then-check verifier, consumes $\ztil$ as a prediction of the future. The robustness of the policy therefore does not entail the robustness of systems that rely on the WAM. The underlying phenomenon is an asymmetry. Corrupting the imagination is easy, since it requires only displacing $\ztil$ from its natural-future manifold. Steering it precisely is hard, since it must reach a specified on-manifold target. We adopt a capability-based threat model with an $\Linf$-bounded observation perturbation. The attacker applies projected gradient descent through the fully differentiable observation-to-imagination map. The same off-manifold property motivates a parameter-free denoiser detector. We evaluate three targets: RynnVLA-002, LingBot-VA, and LaDi-WM. Untargeted corruption is roughly $60\times$ stronger than random and is detected at AUC $1.0$. Targeted control remains bounded. An adaptive attacker evades detection only by forgoing corruption. The reactive policy remains robust to corrupted imagination. A native imagination-driven MPC, however, exhibits the first adversary-specific task failure (at $\epsilon{=}0.01$, success $0.70$ versus $0.05$; Fisher $p<10^{-4}$).
\end{abstract}

\section{Introduction}

Classical vision-language-action (VLA) policies are \emph{reactive}: an observation $o$ is mapped directly to an action $a=\pi(o)$. A second family adopts an \emph{imagine-then-act} design built around a \emph{world-action model} (WAM). The model first imagines a short future $\ztil_{t+1:t+K}$, a latent trajectory of upcoming visual representations. The action is then conditioned on this imagination~\citep{lingbotva,ladiwm}. The imagination provides the policy with a form of internal rehearsal.

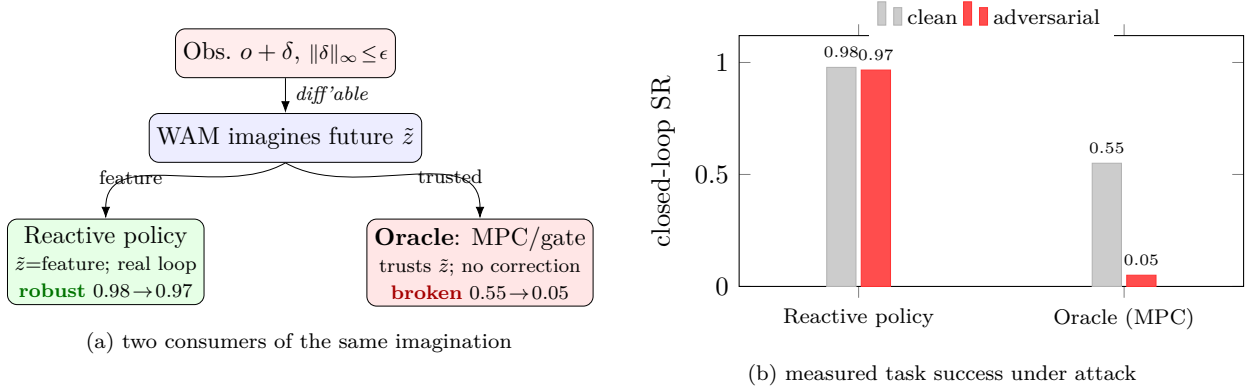
\begin{figure}[t]\centering
\begin{minipage}[c]{0.47\textwidth}\centering
\resizebox{\linewidth}{!}{%
\begin{tikzpicture}[
  font=\small, >=Latex,
  box/.style={draw, rounded corners, align=center, inner sep=3pt, minimum height=7mm},
]
\node[box, fill=red!7] (obs) {Obs.\ $o+\delta$,\ \scriptsize$\|\delta\|_\infty\!\le\!\epsilon$};
\node[box, fill=blue!7, below=5mm of obs] (wam) {WAM imagines future $\ztil$};
\draw[->] (obs) -- (wam) node[midway, right, font=\scriptsize\itshape]{diff'able};
\node[box, fill=green!10, below left=8mm and -8mm of wam, align=center] (pol)
  {Reactive policy\\[-1pt]\scriptsize $\ztil$=feature; real loop\\[-1pt]\scriptsize\textbf{\textcolor{green!45!black}{robust}} $0.98\!\to\!0.97$};
\node[box, fill=red!10, below right=8mm and -8mm of wam, align=center] (orc)
  {\textbf{Oracle}: MPC/gate\\[-1pt]\scriptsize trusts $\ztil$; no correction\\[-1pt]\scriptsize\textbf{\textcolor{red!65!black}{broken}} $0.55\!\to\!0.05$};
\draw[->] (wam.south) to[out=-150,in=90] node[left, font=\scriptsize]{feature} (pol.north);
\draw[->] (wam.south) to[out=-30,in=90] node[right, font=\scriptsize]{trusted} (orc.north);
\end{tikzpicture}}\\[2pt]
{\scriptsize (a) two consumers of the same imagination}
\end{minipage}\hfill
\begin{minipage}[c]{0.50\textwidth}\centering
\begin{tikzpicture}
\begin{axis}[
  ybar, width=\linewidth, height=4.9cm, font=\small,
  ymin=0, ymax=1.12, ylabel={closed-loop SR}, ylabel near ticks,
  symbolic x coords={Reactive policy, Oracle (MPC)}, xtick=data,
  x tick label style={font=\scriptsize, align=center},
  bar width=11pt, enlarge x limits=0.45,
  nodes near coords={\pgfmathprintnumber[fixed,precision=2]{\pgfplotspointmeta}},
  every node near coord/.append style={font=\tiny, anchor=south},
  legend style={font=\scriptsize, at={(0.5,1.16)}, anchor=north, legend columns=2, draw=none},
]
\addplot[fill=gray!40, draw=gray!60] coordinates {(Reactive policy,0.978) (Oracle (MPC),0.55)};
\addplot[fill=red!70, draw=red!80] coordinates {(Reactive policy,0.966) (Oracle (MPC),0.05)};
\legend{clean, adversarial}
\end{axis}
\end{tikzpicture}\\[2pt]
{\scriptsize (b) measured task success under attack}
\end{minipage}
\caption{\textbf{Motivation: policy-robust $\ne$ oracle-robust, with data.} A single $\Linf$-bounded perturbation steers the WAM's imagined future $\ztil$, but its effect depends on who consumes $\ztil$ (a). A \emph{reactive} policy treats $\ztil$ as an internal feature and still executes in a real closed loop, where environment feedback corrects errors. A \emph{downstream oracle} (visual-MPC, safety gate, or verifier) instead trusts $\ztil$ as the future and has no corrective signal. Panel (b) shows the measured contrast. The reactive policy barely moves even at a \emph{large} budget (LingBot-VA, oracle-attacked imagination $\epsilon0.8$: $0.978\!\to\!0.966$; \cref{tab:null}). The oracle MPC collapses at a \emph{tiny} budget (LaDi-WM sampling-MPC $\epsilon0.01$: $0.55\!\to\!0.05$; \cref{tab:mpc}). Robustness of the policy is \emph{not} robustness of systems that trust the WAM.}\label{fig:motivation}
\end{figure}

Most adversarial work on robot policies attacks the \emph{action} pathway and measures action error or task success. We instead ask who consumes the imagination. The imagined future $\ztil$ may serve two very different consumers. The first is a \textbf{reactive policy}, which treats $\ztil$ as an internal feature. Such a policy still executes in a \emph{real closed loop}, where environment feedback continually corrects its errors. The second is a \textbf{downstream oracle}, which treats $\ztil$ as a \emph{trusted prediction of the future} and acts on it without any real-world corrective signal. Examples include a predictive \emph{safety gate}, a \emph{visual MPC} planner, and an \emph{imagine-then-check verifier}. The gate imagines a rollout, predicts success or collision, and decides whether to abort. The planner imagines each candidate action and selects the best-imagined one. This second consumer is the attack surface of our work. When the WAM acts as planner, verifier, or simulator, its imagination is the \emph{sole} information source. No real world is available to correct it, so corrupting the imagination directly corrupts the downstream decision (\cref{fig:motivation}).

\paragraph{Contributions.}
\begin{itemize}\itemsep1pt
\item \textbf{A new attack surface (\cref{sec:method,sec:exp}).} The first white-box integrity attack against the \emph{trusted imagination} of imagine-then-act WAMs. The observation$\to$imagination map is continuous and differentiable (no discrete vector quantization, no in-place cache writes), enabling true projected gradient descent (PGD) on the imagined latent.
\item \textbf{An asymmetry mechanism (\cref{sec:exp,sec:mechanism}).} Imagination is easy to \emph{corrupt} (untargeted) but hard to precisely \emph{steer} (targeted). The two regimes are consistent across targets and are pinned by decoded-frame evidence, not only latent cosine.
\item \textbf{A defense that empirically holds against adaptive attackers (\cref{sec:detect,sec:adaptive}).} Corruption is intrinsically off-manifold; a parameter-free denoiser self-consistency detector reaches AUC $1.0$, and across our adaptive-attacker sweep an attacker cannot evade detection without surrendering corruption.
\item \textbf{An honest task-level null, folded into motivation (\cref{sec:null}).} A reactive policy is robust to corrupted imagination. We turn this into the thesis ``policy robust $\ne$ oracle robust'' and demonstrate, on a native imagination-driven MPC, the first \emph{adversary-specific} real task failure.
\end{itemize}

\section{Related Work}

\paragraph{World models and imagine-then-act policies.}
Several robot policies couple a generative world model with an action head, so that actions are conditioned on an \emph{imagined} future. RynnVLA-002~\citep{rynnvla} unifies action and image generation in one autoregressive backbone. It decodes actions from their own token state, so imagination and action share only the backbone. LingBot-VA~\citep{lingbotva} interleaves video and action tokens in a single causal sequence. An action token attends the just-generated imagined-future latent through a block-causal mask, which makes imagination a genuine input to the action. LaDi-WM~\citep{ladiwm} predicts the \emph{latent} evolution of DINO/SigLIP features with an iterative latent-diffusion world model. It then uses this prediction to refine a diffusion policy's actions. These three targets span the imagination-to-action coupling from disconnected to native (\cref{tab:targets}). Prior work studies such imagination as a way to \emph{help} the policy. We instead treat it as an \emph{attack object}. We also treat it as a signal that a \emph{downstream} planner, verifier, or safety gate may \emph{trust} without a corrective real-world signal.

\paragraph{Adversarial attacks on robot policies.}
Bounded-perturbation attacks~\citep{goodfellow2015,madry2018} and zeroth-order variants~\citep{spall1992} are standard for vision models. They have been applied to the \emph{observation}-to-\emph{action} path of robot policies, where harm is measured as action error or task-success drop. Our threat model differs in its \emph{target}. We attack the imagined future consumed by an oracle, rather than the executed action of a reactive policy. The two can come apart, since an attack that strongly corrupts imagination leaves reactive task success almost unchanged (\cref{sec:null}).

\paragraph{Attacks and robustness of world-action models.}
The closest contemporaneous work attacks world-action models in two ways that differ from ours. JailWAM~\citep{jailwam} is a \emph{black-box, semantic} jailbreak that searches adversarial visual \emph{trajectories} to trigger unsafe behaviors. We are instead white-box and operate within an $\Linf$ pixel budget on a \emph{single} observation, directly through the differentiable observation-to-imagination map. Robustness studies of world-action models~\citep{wamrobustness} measure generalization under \emph{natural} visual and language perturbations, not a worst-case adversary. To our knowledge, no prior work addresses the white-box integrity attack on the \emph{trusted imagination} consumed by a downstream oracle. We pair this attack with a matching off-manifold detector and an adaptive-attacker analysis.

\paragraph{Benchmarks.}
We evaluate on LIBERO~\citep{libero} (long-horizon manipulation); related sim-based evaluation suites such as SimplerEnv~\citep{simplerenv} target real-to-sim transfer and are complementary but out of scope here.

\section{Threat Model}\label{sec:threat}
We adopt a capability-based threat model (\cref{tab:threat}). An imagine-then-act WAM is deployed, and a downstream oracle trusts its imagination $\ztil$ for a decision. The attacker perturbs only the current camera observation, within an $\Linf$ budget. The attacker is white-box in the Kerckhoffs sense and knows the WAM and the encoder. It does \emph{not} modify weights, does \emph{not} touch the oracle, and has \emph{no} access to the true future. Harm is measured at the \emph{decision level} of the oracle, through verifier flip rate, MPC mis-selection, and divergence of the trusted imagination. Task-level closed-loop success is reported as the reactive-robustness control. Among the three oracle instantiations, only the visual-MPC consumer is instantiated end-to-end with closed-loop success (LaDi-WM, \cref{sec:mpc}). The imagine-then-check verifier (\cref{sec:untargeted}) and the predictive safety gate (\cref{sec:limits}) are evaluated as decision-level probes, whose scope we state explicitly.

\begin{table}[h]\centering\small
\caption{Capability-based threat model.}\label{tab:threat}
\begin{tabular}{@{}l p{0.72\textwidth}@{}}
\toprule
& Setting\\
\midrule
Deployment & Imagine-then-act WAM; imagination $\ztil$ trusted by a downstream oracle (gate / MPC / verifier).\\
Attacker capability & $\Linf$-bounded perturbation of the observation, $\|\delta\|_\infty\le\epsilon$; white-box.\\
Out of scope for attacker & no weight modification; no access to the oracle internals; no access to the true future.\\
Objective & untargeted: maximize divergence of $\ztil$ from the model's own clean imagination; or targeted: steer $\ztil$ to an attacker-chosen scene.\\
Harm metric & decision-level (verifier flip, MPC mis-selection, trusted-imagination divergence); plus task SR as control.\\
\bottomrule
\end{tabular}
\end{table}

\section{Method}\label{sec:method}
\begin{figure}[t]\centering
\begin{tikzpicture}[
  font=\small, >=Latex,
  box/.style={draw, rounded corners, align=center, inner sep=4pt, minimum height=9mm},
  frozen/.style={box, fill=blue!7},
]
\node[box, fill=red!7] (obs) {$o+\delta$\\[-1pt]\scriptsize observation};
\node[frozen, right=9mm of obs] (enc) {$E$\\[-1pt]\scriptsize encoder};
\node[frozen, right=8mm of enc] (wm) {$W$\\[-1pt]\scriptsize world model};
\node[box, fill=blue!4, right=8mm of wm] (z) {$\ztil_{\mathrm{adv}}$\\[-1pt]\scriptsize imagined future};
\node[box, fill=gray!8, right=9mm of z] (orc) {oracle $\mathcal{V}$\\[-1pt]\scriptsize decision};
\draw[->] (obs)--(enc); \draw[->] (enc)--(wm); \draw[->] (wm)--(z); \draw[->] (z)--(orc);
\node[font=\scriptsize\itshape, below=2mm of wm, text=blue!50!black] (diff) {no VQ, no in-place cache $\Rightarrow$ end-to-end differentiable};
\draw[->, dashed, red!70!black] (z.south) to[out=-90,in=-90] node[below, font=\scriptsize, pos=0.5]{PGD\ $\nabla_\delta L$} (obs.south);
\node[box, fill=yellow!12, above=7mm of z, align=left, font=\scriptsize] (obj)
  {\textbf{Objective} (\cref{def:obj}):\\
   untargeted: push $\ztil$ \emph{off-manifold}, away from $\zcln$\\
   targeted: steer $\ztil$ toward attacker's $\ztgt$};
\draw[->, dotted] (obj.south) -- (z.north);
\node[box, fill=green!10, below right=7mm and 1mm of orc, align=center, font=\scriptsize] (det)
  {\textbf{Detector} (\cref{sec:detect}):\\ denoiser self-consistency\\ $s(\ztil)\gtrless\tau$ (no clean ref.)};
\draw[->] (z.east) to[out=-20,in=180] (det.west);
\end{tikzpicture}
\caption{\textbf{Method overview.} The attacker adds an $\Linf$-bounded $\delta$ to the observation and back-propagates through the frozen encoder $E$ and world model $W$ to run true PGD on the imagined future $\ztil$. These two modules form a continuous, fully differentiable observation$\to$imagination map, with no discrete VQ and no in-place KV-cache writes. There are two objectives (\cref{def:obj}): \emph{untargeted} corruption pushes $\ztil$ off the natural-future manifold (easy, about $60\times$ random), and \emph{targeted} steering aims $\ztil$ at an attacker-chosen $\ztgt$ (bounded). A downstream oracle $\mathcal{V}$ consumes $\ztil$, and a parameter-free denoiser self-consistency detector flags off-manifold corruption before the oracle acts.}\label{fig:method}
\end{figure}
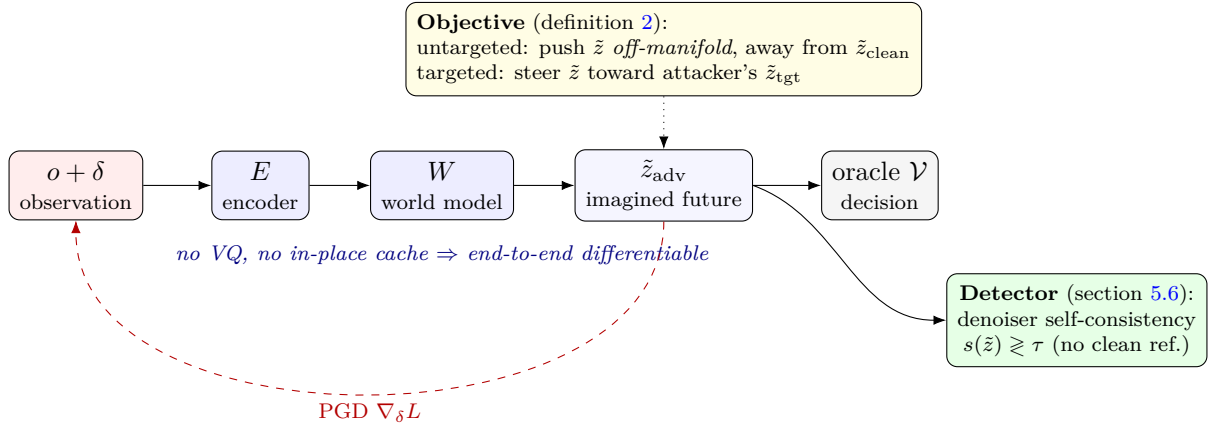

A WAM comprises an encoder $E$ and a world model $W$. The observation is encoded, and the world model generates the imagined future $\ztil=W(E(o),c)$ under instruction $c$. A downstream oracle $\mathcal{V}$ then makes a decision from $\ztil$. The attacker optimizes $\delta$ to change $\ztil$ (\cref{fig:method}).

\begin{definition}[Adversarial perturbation of imagination]\label{def:pert}
The attacker forms $o'=o+\delta$ with $\|\delta\|_\infty\le\epsilon$, which induces the corrupted imagination $\zadv=W(E(o+\delta),c)$. For continuous diffusion-latent WAMs the map $o\mapsto\ztil$ is end-to-end differentiable, with no discrete VQ and no in-place cache writes. True gradient PGD is therefore applicable.
\end{definition}

\paragraph{Two objectives: corrupt vs.\ steer.}
\begin{definition}[Untargeted corruption / targeted steering]\label{def:obj}
\emph{Untargeted} maximizes divergence from the clean imagination,
\begin{equation}
L_{\mathrm{untgt}}(\delta)=-\big(1-\cos(\zadv,\zcln)\big),\qquad
\text{corruption}=1-\cos(\zadv,\zcln).
\end{equation}
\emph{Targeted} drives the imagination toward an attacker-chosen $\ztgt$ (the imagination for a swapped instruction or scene),
\begin{equation}
L_{\mathrm{tgt}}(\delta)=\|\zadv-\ztgt\|^2,\qquad
\mathrm{gap\_closed}=1-\frac{\|\zadv-\ztgt\|}{\|\zcln-\ztgt\|}.
\end{equation}
$\mathrm{gap\_closed}{=}1$ means the imagination is perfectly steered to the target; ${=}0$ means unchanged; ${<}0$ means farther than clean.
\end{definition}

\paragraph{Attack and detector.}
\Cref{alg:attack} gives the PGD attack and the off-manifold detector. One implementation detail matters. All $\texttt{\_reset}(c)$ calls must be wrapped in \texttt{no\_grad}, and \emph{all} parameters (transformer, text encoder, VAE) must be frozen. Otherwise an un-frozen text encoder caches a graph-bearing text context that is reused across forward passes, which triggers a ``backward through the graph a second time'' error. The gradient is sometimes unavailable, for example when actions read $\ztil$ through an \emph{in-place} KV-cache write that severs the gradient and yields $\|g\|{=}0$. In that case we fall back to SPSA~\citep{spall1992}, which we verified converges.

\begin{algorithm}[h]
\caption{Imagination attack (true PGD) + off-manifold detector}\label{alg:attack}
\begin{algorithmic}[1]
\Require WAM $W$, encoder $E$, observation $o$, instruction $c$, budget $\epsilon$, steps $T$, step size $\alpha$
\State Freeze \textbf{all} parameters; wrap every $\texttt{\_reset}(c)$ in \texttt{no\_grad}
\State $\delta\gets 0$
\For{$t=1\ldots T$}
  \State $\zadv\gets W(E(o+\delta),c)$;\quad $g\gets\nabla_\delta L(\delta)$ \Comment{$L$ = untargeted or targeted (\cref{def:obj})}
  \State $\delta\gets\Pi_{\|\cdot\|_\infty\le\epsilon}\big(\delta-\alpha\,\mathrm{sign}(g)\big)$
\EndFor
\State \Return $\delta$
\Statex \textbf{Detector (no clean reference):}
\State $s(\ztil)\gets$ mean future-frame velocity-prediction norm at small noise levels $t\in\{0.8,0.9,0.97\}$
\State \textbf{if} $s(\ztil)>\tau$ \textbf{then} flag $\ztil$ as corrupted
\end{algorithmic}
\end{algorithm}

\section{Experiments}\label{sec:exp}
All experiments are run on a single-A100 cluster (Adelaide Phoenix). The benchmark is LIBERO-10 (long horizon). Closed-loop rollouts use $n{=}20$ with Wilson $95\%$ intervals and Fisher's exact test.

\subsection{Setup: three targets spanning the imagination-action coupling}
We study three imagine-then-act targets chosen to span the strength of the imagination$\to$action coupling (\cref{tab:targets}).

\paragraph{Scale at a glance (read with the caption of each table).} Latent-level attack and detection probes are reported over $N$ tasks. These are targeted steering at $N{=}1$ to $8$ (\cref{tab:targeted}), untargeted corruption and the negative control at $N{=}5$ (\cref{tab:untargeted,tab:c1}), detection at $N{=}8$, and the adaptive attacker at $N{=}4$ (\cref{tab:adaptive}). The closed-loop results use full episodes with Wilson $95\%$ intervals. The reactive null aggregates $143$ to $190$ episodes per condition across all $10$ tasks (\cref{tab:null}). The LaDi-WM MPC failure is, at present, a single task (task~4) with $N{=}20$ per $\epsilon$ (\cref{tab:mpc,tab:isolation}). We flag this single-task scope as a limitation (\cref{sec:limits}) rather than smoothing over it. The conclusions that do \emph{not} depend on rollouts (untargeted corruption, detection, adaptive defense) are the most robust.

\begin{table}[h]\centering\small
\caption{Three imagine-then-act targets.}\label{tab:targets}
\begin{tabular}{@{}l l p{0.30\textwidth} p{0.22\textwidth}@{}}
\toprule
Target & Imagination & Imagination$\to$action coupling & Role\\
\midrule
RynnVLA-002 & discrete VQ tokens & \emph{disconnected} (action reads its own token hidden) & negative control / zero transfer\\
LingBot-VA & continuous diffusion latent $\ztil$ & same-model, weak at task level & oracle attack/defense (O1--O8)\\
LaDi-WM & latent diffusion, native refine & strong (action-conditioned dynamics) & real task failure + MPC\\
\bottomrule
\end{tabular}
\end{table}

\subsection{RynnVLA-002: coupling exists but transfers nothing at task level}
On RynnVLA-002 the action head reads its own token hidden state and \emph{does not consume the imagination at inference}. Imagination is a separate generation path, and the two heads share only the backbone. We find a strong but undirected same-model gradient coupling ($|\cos(g_a,g_{\mathrm{img}})|{=}0.31$, $534\times$ random). Same-model transduction is real but weaker than directly attacking the action, with about $20\%$ catastrophic state reversals under an impact proxy. A \emph{cross-model} rollout, however, shows \textbf{zero transfer}. Imagining with one checkpoint and transferring the perturbation to a victim yields task success $=100\%=$ random $=$ clean, whereas directly attacking the victim's action yields $0\%$. An early oracle-deception probe already hints at the central asymmetry: an untargeted push corrupts ($1.6\times$ random), but targeted steering is unstable. This result motivates the move to continuous, differentiable targets and to the oracle threat model.

\subsection{Targeted manipulation is bounded}\label{sec:targeted}
On LingBot-VA true PGD is available, since the $o\to\ztil$ map is differentiable ($\|g\|\!\sim\!0.094$). Targeted steering is \emph{adversarial-direction-specific}, as random achieves $\approx0$. It is nonetheless \emph{bounded} once the target is a genuinely different scene (\cref{tab:targeted}). When the target is the same observation with a swapped instruction (already $89\%$ similar), $\mathrm{gap\_closed}$ reaches $0.45$ to $0.52$ and a verifier is fooled $5/5$. For a cross-scene target the latent cosine moves to $\mathrm{gap\_closed}{=}0.207$, but the decoded frames and the verifier do not. The verifier flips $0/8$, and the decoded final frames remain the original scene. Directing imagination precisely to an arbitrary different scene is therefore bounded.

\begin{table}[h]\centering\small
\caption{Targeted steering (LingBot-VA): adversarial-specific but bounded for cross-scene targets.}\label{tab:targeted}
\begin{tabular}{@{}l l l l l@{}}
\toprule
Exp & Setting & adv.\ gap\_closed & random & verifier / decode\\
\midrule
O1 & same-obs swap-instr., $N{=}1$ & $0.447$ & $-0.079$ & $\cos(\mathrm{adv},\mathrm{tgt})\,0.891{\to}0.966$\\
O2 & $\epsilon\,0.1$--$0.4$ sweep, single obs & $0.46$--$0.52$ & all neg. & verifier fooled $5/5$ vs.\ random $1/5$\\
O3 & \textbf{cross-scene}, $N{=}8$, $\epsilon0.2$ & $0.207\pm0.056$ & $-0.002\pm0.008$ & \textbf{verifier $0/8$; decode $\approx$ clean}\\
\bottomrule
\end{tabular}
\end{table}

\begin{figure}[h]\centering
\setlength{\tabcolsep}{3pt}
\begin{tabular}{c@{\ }c@{\ }c@{\ }c@{\ }c}
\includegraphics[width=0.27\textwidth]{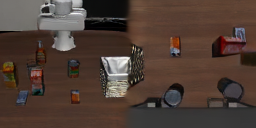} &
\raisebox{0.7\height}{\large$\approx$} &
\includegraphics[width=0.27\textwidth]{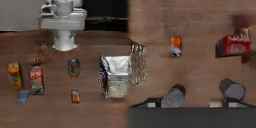} &
\raisebox{0.6\height}{\large$\neq$} &
\includegraphics[width=0.27\textwidth]{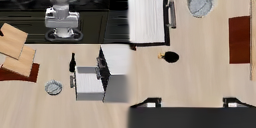}\\
(a) clean imagination & & (b) targeted adversarial & & (c) attacker's target\\
\end{tabular}
\caption{Targeted steering is bounded (LingBot-VA, decoded imagined frame $f{=}12$, task~$0\!\to\!3$; each panel shows the two decoded camera views, \emph{left}: third-person agentview, \emph{right}: wrist). Despite a latent-cosine $\mathrm{gap\_closed}{=}0.207$, the adversarial imagination (b) still depicts the \emph{original} scene (the basket-and-cans tabletop of (a), with only mild artifacts; $\mathrm{b}\!\approx\!\mathrm{a}$) and does \emph{not} become the attacker's target scene (c, a different kitchen layout; $\mathrm{b}\!\neq\!\mathrm{c}$). Latent cosine moves; pixels and the verifier do not.}\label{fig:targeted}
\end{figure}

\subsection{Untargeted corruption is easy and perceptually visible}\label{sec:untargeted}
This is the strongest result on the attack side. We run untargeted PGD that maximizes divergence from the true imagination, over $N{=}5$ tasks (\cref{tab:untargeted}). A fidelity verifier flips to ``imagination not trustworthy'' when $\cos<\tau$. Adversarial corruption exceeds random by up to $60\times$. It flips the verifier at $100\%$ from $\epsilon{=}0.1$, with $0\%$ false positives. Decoding confirms \emph{perceptually visible} damage, including object smearing, ghosting, and artifacts. This contrasts with the perceptually invisible targeted case (\cref{sec:targeted}).

\begin{table}[h]\centering\small
\caption{Untargeted corruption $\times$ fidelity verifier (LingBot-VA), $N{=}5$ tasks.}\label{tab:untargeted}
\begin{tabular}{lllll}
\toprule
$\epsilon$ & adv.\ corruption $1{-}\cos$ & random & ratio & verifier flip (adv / rand)\\
\midrule
$0.05$ & $0.135\pm0.05$ & $0.002$ & $\sim60\times$ & n/a\\
$0.1$  & $0.197\pm0.056$ & $0.006$ & $\sim30\times$ & $@\tau0.9$: $100\%/0\%$\\
$0.2$  & $0.266\pm0.03$ & $0.013$ & $\sim20\times$ & $@\tau0.8$: $100\%/0\%$\\
$0.8$  & $0.467\pm0.09$ & $0.109$ & $\sim4\times$  & $100\%/$n/a\\
\bottomrule
\end{tabular}
\end{table}

\begin{figure}[h]\centering
\setlength{\tabcolsep}{3pt}
\begin{tabular}{c@{\ }c@{\ }c}
\includegraphics[width=0.40\textwidth]{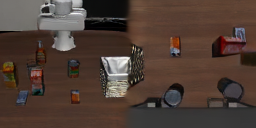} &
\raisebox{0.7\height}{\large$\Rightarrow$} &
\includegraphics[width=0.40\textwidth]{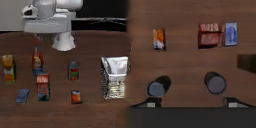}\\
(a) clean imagination & & (b) untargeted adversarial ($\epsilon0.2$)\\
\end{tabular}
\caption{Untargeted corruption is perceptually visible (LingBot-VA, decoded imagined frame $f{=}12$, task~$0$; each panel shows the two decoded camera views, \emph{left}: third-person agentview, \emph{right}: wrist). Compared to the clean imagined future (a), the corrupted imagination (b) exhibits object smearing, ghosting, and a torn region on the right. In direct contrast to the targeted case (\cref{fig:targeted}), the damage here is plainly visible to any consumer of the imagined frames.}\label{fig:untargeted}
\end{figure}

\paragraph{Which verifier, honestly?} The fidelity verifier above thresholds $\cos(\zadv,\zcln)$, the very quantity the untargeted attack minimizes. Its $100\%$ flip is therefore partly by construction, and it should be read as a \emph{trust/fidelity} gate rather than an independent oracle. We also evaluate an \emph{independent} decision-level verifier (V1), a nearest-prototype task classifier over clean imagined futures, with clean accuracy $1.000$. Under untargeted corruption it misclassifies the imagined task in only $12\%$ of cases, against $0\%$ for random (targeted false-accept-as-target $0\%$, $N{=}8$, $\epsilon0.2$). The most faithful decision-level harm is therefore the \emph{availability} attack of \cref{sec:limits}. A real denoiser safety gate, when attacked, drops effective task success by $0.18$ to $0.39$. We report all three measures rather than only the largest.

\subsection{Negative control: corruption is imagination-channel-specific}\label{sec:c1}
Is the corruption a \emph{directed} exploit of the imagination channel, or merely generic visual damage that any consumer would see? For the same bounded observation perturbation, we compare the divergence on two quantities. The first is the imagined future frames $\tilde z_{[1:]}$. The second is the current-observation read-out latent (frame $0$), which is what the no-imagination reactive consumer sees. The amplification $\mathrm{amp}=\mathrm{div}_{\mathrm{future}}/\mathrm{div}_{\mathrm{obs}}$ is $24.5\times$ at $\epsilon0.1$ and stays above $1$ throughout (\cref{tab:c1}). Moreover $\mathrm{div}_{\mathrm{future}}(\mathrm{adv})\gg\mathrm{div}_{\mathrm{future}}(\mathrm{rnd})$. The corruption is thus a directed exploit of the imagination channel and is largely invisible to an observation-faithful consumer. That channel is exactly the one a downstream oracle uniquely trusts.

\begin{table}[h]\centering\small
\caption{Negative control (LingBot-VA, $N{=}5$): corruption amplifies on the imagination channel relative to the obs read-out.}\label{tab:c1}
\begin{tabular}{lllll}
\toprule
$\epsilon$ & $\mathrm{div}_{\mathrm{future}}$ (adv) & $\mathrm{div}_{\mathrm{obs}}$ (adv) & amp & $\mathrm{div}_{\mathrm{future}}$ (rnd)\\
\midrule
$0.1$ & $0.198\pm0.045$ & $0.0083$ & $24.5\times$ & $0.004$\\
$0.2$ & $0.282\pm0.029$ & $0.0325$ & $8.7\times$  & $0.012$\\
$0.4$ & $0.378\pm0.042$ & $0.1044$ & $3.6\times$  & $0.040$\\
$0.8$ & $0.545\pm0.043$ & $0.2656$ & $2.1\times$  & $0.110$\\
\bottomrule
\end{tabular}
\end{table}

\subsection{Detection: denoiser self-consistency reaches AUC $1.0$}\label{sec:detect}
Corrupted imagination departs the model's own learned manifold, and is thus \emph{off-manifold}. Our detector is parameter-free: it is the mean future-frame velocity-prediction norm at small noise levels, and it needs \emph{no} clean reference. This detector separates clean from adversarial almost perfectly (\cref{tab:detect}). Surface statistics such as latent RMS and max-abs are at chance level, since the attack stays within the $\Linf$ budget and does not inflate norms.

\begin{table}[h]\centering\small
\caption{Detection AUC (clean vs.\ adversarial), LingBot-VA.}\label{tab:detect}
\begin{tabular}{lllll}
\toprule
Detector & $\epsilon0.1$ & $\epsilon0.2$ & $\epsilon0.4$ & TPR@5\%FPR\\
\midrule
surface RMS / max-abs & $\sim0.5$ & $\sim0.5$ & $\sim0.5$ & chance\\
simple clip-corruption & $0.5$--$0.62$ & n/a & $0.86$ & weak at low $\epsilon$\\
\textbf{denoiser self-consistency} & $0.89$ & $1.00$ & $1.00$ & $62\%/100\%/100\%$\\
\bottomrule
\end{tabular}
\end{table}

\subsection{Adaptive attacker: the defense holds}\label{sec:adaptive}
A reviewer will ask whether a \emph{knowledgeable} attacker can evade the detector. We let the attacker minimize $L=-\text{corruption}+\lambda\cdot\text{det\_score}$ (\cref{tab:adaptive}). As $\lambda$ grows, AUC falls below $0.5$ and the det-score approaches the clean baseline $0.369$. At the same time the corruption falls with it, from $0.279$ to $0.064$. Being off-manifold is part of what it means to corrupt the imagination. To evade detection the attacker must stay on the manifold, which means not corrupting. The two goals cannot both be met, so the defense holds against adaptive attackers.

\begin{table}[h]\centering\small
\caption{Adaptive attacker vs.\ the denoiser detector (LingBot-VA). Evasion requires surrendering corruption.}\label{tab:adaptive}
\begin{tabular}{llll}
\toprule
$\lambda$ & corruption $1{-}\cos$ & det\_score & AUC\\
\midrule
$0.0$ & $0.279$ & $0.562$ & $1.00$\\
$0.5$ & $0.241$ & $0.407$ & $0.88$\\
$2.0$ & $0.156$ & $0.345$ & $0.31$\\
$8.0$ & $0.064$ & $0.337$ & $0.06$\\
\bottomrule
\end{tabular}
\end{table}

\subsection{Universal transfer}\label{sec:universal}
A single scene-agnostic $\delta$ ($|\delta|{=}21.25$) is optimized on train tasks $[0,3]$ and then frozen. It still corrupts the held-out tasks $[4,7]$, with test corruption $0.199$ against random $0.039$ ($\sim5\times$; train $0.356$ against $0.034$). This indicates a deployable, scene-agnostic threat.

\subsection{LaDi-WM sampling MPC: adversary-specific real task failure}\label{sec:mpc}
\begin{figure}[h]\centering
\begin{tikzpicture}
\begin{axis}[
  width=0.9\textwidth, height=7cm, font=\small,
  xlabel={$\Linf$ perturbation budget $\epsilon$}, ylabel={closed-loop success rate},
  xmin=-0.002, xmax=0.07, ymin=0, ymax=0.85,
  xtick={0,0.01,0.03,0.06}, scaled x ticks=false,
  xticklabels={$0$,$0.01$,$0.03$,$0.06$},
  ytick={0,0.2,0.4,0.55,0.7,0.8},
  legend pos=north east, legend cell align=left, legend style={font=\small},
  grid=both, grid style={gray!15},
]
\fill[red!12] (axis cs:0.005,0) rectangle (axis cs:0.04,0.85);
\node[red!60!black, font=\scriptsize\bfseries, align=center] at (axis cs:0.0225,0.79)
  {adversary-specific\\[-2pt]window};
\addplot[densely dashed, gray, thick] coordinates {(-0.002,0.55) (0.07,0.55)};
\addlegendentry{clean baseline ($0.55$)}
\addplot[green!55!black, mark=*, mark size=3pt, very thick,
  nodes near coords, every node near coord/.append style={font=\scriptsize, anchor=south, yshift=1pt}]
  coordinates {(0,0.55) (0.01,0.70) (0.03,0.65) (0.06,0.40)};
\addlegendentry{random noise}
\addplot[red!80!black, mark=square*, mark size=3pt, very thick,
  nodes near coords, every node near coord/.append style={font=\scriptsize, anchor=north, yshift=-2pt}]
  coordinates {(0,0.55) (0.01,0.05) (0.03,0.05) (0.06,0.0)};
\addlegendentry{adversarial}
\draw[<->, blue!60!black, thick] (axis cs:0.0115,0.68) -- (axis cs:0.0115,0.07);
\node[blue!60!black, font=\scriptsize, anchor=west, align=left] at (axis cs:0.013,0.37)
  {$0.70$ vs $0.05$\\[-2pt]Fisher $p\!<\!10^{-4}$};
\end{axis}
\end{tikzpicture}
\caption{\textbf{LaDi-WM MPC closed-loop SR vs.\ perturbation budget} (task~4, $N{=}20$ per point, Wilson $95\%$). In the shaded \emph{adversary-specific window} ($\epsilon\!\in\![0.01,0.03]$), random noise of the same magnitude is harmless ($\mathrm{SR}\!\approx\!0.65$ to $0.70\!\approx$ clean) while the adversarial direction \emph{collapses} success to $0.05$ (non-overlapping intervals, Fisher $p\!<\!10^{-4}$). At the large budget $\epsilon0.06$ both collapse (saturation): magnitude alone, not adversariality, dominates there.}\label{fig:sweep}
\end{figure}
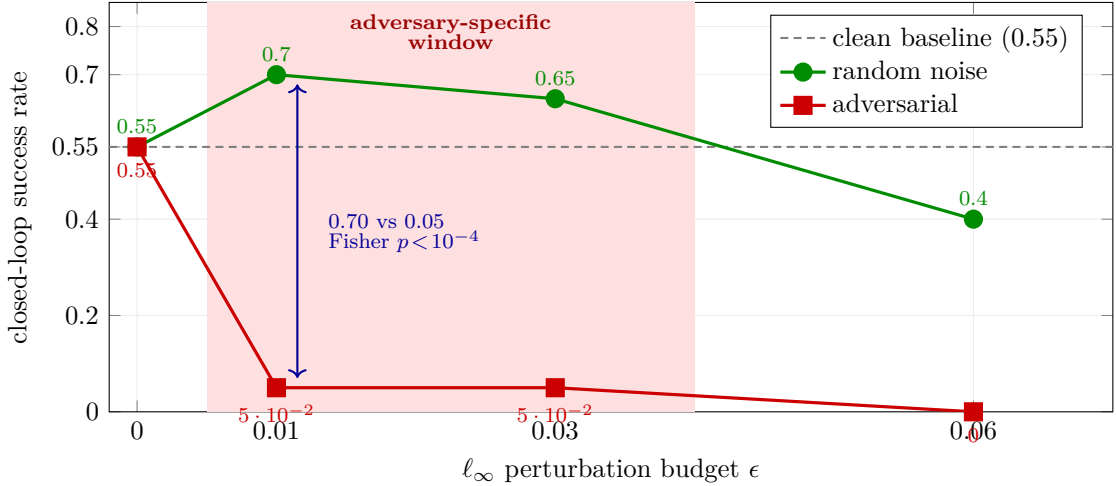

Unlike the reactive null (\cref{sec:null}), this setting produces genuine task failure. LaDi-WM's policy \emph{natively} refines actions with imagination. We build a sampling MPC on top of it: every $5$ steps it samples $K{=}4$ candidate actions, imagines and scores each, and takes the $\argmax$. The attacker corrupts imagination via $\Linf$-PGD on the observation. We evaluate on task~4 with $N{=}20$ (\cref{fig:sweep,tab:mpc}).

\begin{table}[h]\centering\small
\caption{LaDi-WM MPC: SR-vs-$\epsilon$ (task~4, $N{=}20$). Clean $=0.55$.}\label{tab:mpc}
\begin{tabular}{llll}
\toprule
$\epsilon$ & random SR & adv.\ SR & reading\\
\midrule
$0.01$ & $14/20=0.70$ & $1/20=0.05$ & random $\approx$ clean, adversarial collapses = adversary-specific\\
$0.03$ & $13/20=0.65$ & $1/20=0.05$ & same direction, adversary-specific window\\
$0.06$ & $8/20=0.40$  & $0/20=0.0$   & saturation: random also collapses\\
\bottomrule
\end{tabular}
\end{table}

At $\epsilon{=}0.01$ the random Wilson interval $[0.48,0.85]$ and the adversarial interval $[0.01,0.24]$ do not overlap (Fisher $p<10^{-4}$). Random noise of the same magnitude is harmless ($0.70\approx$ clean), whereas the adversarial direction collapses success to $0.05$. We also note an honest self-correction. An earlier single-point read at $\epsilon{=}0.06$, where random and adversarial both collapse, was mistaken for ``mere visual fragility''. The full sweep shows that this point is the \emph{saturation} regime. The adversary-specific window genuinely exists at $\epsilon\in[0.01,0.03]$. There, both $\epsilon{=}0.01$ and $\epsilon{=}0.03$ show adversarial SR $=1/20=0.05$ against random $\ge0.65$ at full $N{=}20$.

\paragraph{Causal isolation.} A five-condition probe at $\epsilon{=}0.06$ and $N{=}20$ localizes the harm (\cref{tab:isolation}). In the \texttt{score\_only} condition we feed the adversarial observation only to the MPC scoring step and execute on a clean observation. Success then \emph{recovers} to $0.60$, so the explicit outer planning and ranking layer is innocent. The \texttt{final\_only} result of $0$ does \emph{not}, however, license a ``pure direct-vision'' claim. The execution step's direct-vision path and the in-\texttt{act} imagination-refinement path share the \emph{same} corrupted image and the \emph{same} DINO-SigLIP encoder. An observation-space attacker cannot separate these two paths. This is why the oracle threat model is needed. Only when the attacker manipulates the \emph{predicted latent} directly, in the trusted downstream z-space, does ``imagination'' become a surface that can be attacked and defended on its own.

\begin{table}[h]\centering\small
\caption{Causal isolation on LaDi-WM MPC ($\epsilon0.06$, $N{=}20$).}\label{tab:isolation}
\begin{tabular}{lll}
\toprule
Condition & SR & meaning\\
\midrule
clean & $11/20=0.55$ & baseline\\
random (matched magnitude) & $8/20=0.40$ & random control\\
adversarial (both paths) & $0/20=0.0$ & full attack\\
\texttt{score\_only} (adv.\ scoring, clean execution) & $12/20=0.60$ & outer planner \emph{innocent} (recovers)\\
\texttt{final\_only} (clean scoring, adv.\ execution) & $0/20=0.0$ & execution-step vision is the carrier\\
\bottomrule
\end{tabular}
\end{table}

\section{Mechanism: off-manifold is intrinsic to corrupting imagination}\label{sec:mechanism}
A single mechanism accounts for the positive results. Imagination $\ztil$ lives on a low-dimensional manifold $\mathcal{M}$ of natural futures learned by the model. Untargeted corruption amounts to pushing $\ztil$ off $\mathcal{M}$. The target set is huge, since any ``wrong'' direction will do, so corruption is easy ($60\times$ random) and \emph{necessarily} off-manifold. The denoiser prior therefore flags it (AUC $1.0$). Targeted steering must instead land $\ztil$ at another specific point of $\mathcal{M}$. It must both leave the original point and re-land on a particular on-manifold location, which is a strong constraint, so it is bounded (gap $0.2$, decode unchanged, verifier $0/8$). Evading detection means staying on $\mathcal{M}$, which means not corrupting, so AUC and corruption collapse together (\cref{sec:adaptive}). The same mechanism explains why corruption is easy, why steering is hard, and why corruption is detectable.

\section{The task-level null, and why it motivates the oracle threat}\label{sec:null}
We state a negative result that constrains the rest of the paper. A moved imagination cosine, or a flip of our own verifier, is not the same as real task failure. In a \emph{reactive} closed loop, corrupted imagination has essentially no task-level effect (\cref{tab:null}). Clean success is $0.978$, oracle-attacked imagination at $\epsilon0.8$ is still $0.966$, and a direct observation attack at $\epsilon1.5$ leaves $0.943$. At mid-budget a universal adversarial $\delta$ is in fact \emph{more harmless} than random noise: the random noise drives SR to $0$, whereas the adversarial $\delta$ preserves $0.83$ to $1.0$.

\begin{table}[h]\centering\small
\caption{Reactive task-level success: the honest null (aggregated rollouts).}\label{tab:null}
\begin{tabular}{lll}
\toprule
Condition & real SR & reading\\
\midrule
clean & $0.978$ & baseline\\
oracle-attacked imagination $\epsilon0.8$ & $0.966$ & barely drops\\
direct observation attack $\epsilon1.5$ & $0.943$ & also unmoved\\
universal adversarial $\epsilon0.8$--$1.2$ & $0.83$--$1.0$ (vs.\ random$\to0$) & adversarial \emph{more harmless} than random\\
\bottomrule
\end{tabular}
\end{table}

This null is the motivation rather than a failure: \emph{policy robustness $\ne$ robustness of a downstream verifier or planner that trusts the WAM}. The same ``policy is robust'' data is a negative result for the reactive consumer and an attack surface for the oracle consumer. LaDi-WM's MPC (\cref{sec:mpc}) is exactly an oracle consumer that natively plans with imagination, and the adversary-specific real failure appears there.

\section{Limitations}\label{sec:limits}
Several limitations remain. First, under observation-space attacks the direct-vision and imagination-refinement paths share an encoder and cannot be separated. Imagination-specific attribution is therefore only possible under the oracle threat model, with direct z-space manipulation (\cref{sec:mpc}). Second, targeted steering is bounded: precisely directing imagination to an arbitrary scene is not achievable (gap $0.2$, decode unchanged, verifier $0/8$). The core threat is \emph{corruption} rather than arbitrary control. Third, the availability claim is downgraded. An early $N{=}3$ pilot suggested a catastrophic effective-SR collapse and gate AUC $1.00$. At full scale ($N{=}179$) it is instead a \emph{mild but statistically significant} availability loss, with an effective-SR drop of $-0.18$ to $-0.39$ across FPR and gate AUC $0.771$, which is the number we report. The hardest contributions (untargeted corruption, detection, adaptive-attacker defense) do not depend on rollouts. Fourth, the evaluation is simulation only and limited in closed-loop scale. The MPC rollouts are currently task~4 with $N{=}20$ across an $\epsilon$ sweep, and a multi-task study with $20$ to $50$ episodes is needed before any universality claim. A semantic CLIP verifier is unreliable at the decoded resolution, so only the denoiser fidelity verifier is used as primary evidence.

\section{Conclusion}
WAM imagination is easy to corrupt and hard to steer precisely. Untargeted corruption flips a verifier at $\epsilon0.1$, is perceptually visible, and is detectable at AUC $1.0$, whereas targeted steering stays bounded. This asymmetry is dangerous for any oracle that trusts a WAM's predicted future, whether a planner, a verifier, or a simulator. The reactive task-level null, in which the policy stays robust, becomes the central motivation. A native imagination-driven MPC then shows the first adversary-specific real task failure. Because corruption is off-manifold, the resulting detector holds even against adaptive attackers.

\paragraph{Reproducibility.} All numbers are real, measured on a single-A100 cluster. Code, attack/defense scripts, decoded frames, and per-condition rollout JSONs are available from the authors.

\small
\nocite{idr2025,duba2024,ebba2024,tripleba2025,skillreducer2026,resourcereservation2025,attrcloak2025,mimir2025}
\bibliographystyle{plainnat}
\bibliography{refs}

\end{document}